# Dual-Precision Deep Neural Network

Jae Hyun Park
Department of AI
Sungkyunkwan University
Suwon, South Korea
xoxc4565@g.skku.edu

Ji Sub Choi
College of Information and
Communication Engineering
Sungkyunkwan University
Suwon, South Korea
cjs3450@skku.edu

Jong Hwan Ko*
College of Information and
Communication Engineering
Sungkyunkwan University
Suwon, South Korea
jhko@skku.edu

## ABSTRACT
On-line Precision scalability of the deep neural networks(DNNs) is a critical feature to support accuracy and complexity trade-off during the DNN inference. In this paper, we propose dual-precision DNN that includes two different precision modes in a single model, thereby supporting an on-line precision switch without re-training. The proposed two-phase training process optimizes both low- and high-precision modes.

## 1. INTRODUCTION
Deep Neural Networks (DNNs) require a large number of parameters and operations to be stored and computed, which act as a critical challenge of accelerating DNN inference on the edge devices with limited resources. One of the solutions for this challenge can be utilizing the quantization method that reduces the numerical precision of the weight parameters and/or activations, which determines the accuracy and required compute/memory resources of the inference. As the accuracy requirement and available resources can vary due to the nature of dynamic environment of the edge device applications, a critical feature for DNN quantization is on-line precision scalability of DNN models, supporting dynamic trade-off between accuracy and complexity of the DNN inference.

One solution for this demand is to store two separate DNN models trained for different target precision ($b$-bit and $(b+1)$-bit precision), and apply either of the two depending on the accuracy/complexity requirements (Fig. 1(a)). However, this solution requires additional memory for storing the two DNN models. Another solution can be transforming a DNN model on-demand for the specific target precision as in Fig. 1(b). However, the existing precision scaling methods generally require retraining of the DNN models, which limits an on-line switch of the model precision. Although post quantization techniques can be applied to remove the need for re-training, the accuracy cannot be guaranteed when down-scaling into the low precision.

In this paper, we propose a new method to train dual-precision DNNs that can switch between the two precision modes without the need for re-training. By sharing the common bits ($b$ bits) between the low- and high-precision weights, the dual-precision DNN allows dynamic on-line switch between the two precision modes only with truncation (precision down-scaling) or concatenation (precision up-scaling) of the last 1 bits. To optimize both precision modes in a single model, we propose the training process with two phases - first phase for shared-bit training and second phase for the entire high-precision training. After applying the proposed training method, the first $b$ bits and the entire $b+1$ bits of the weights are optimized for the low-precision and high-precision modes, respectively, each showing comparable accuracy to the baseline models trained for the target precision using the existing quantization method.

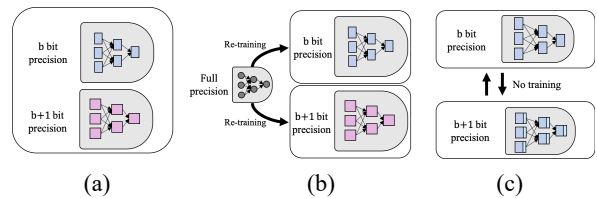

Figure 1. Solutions for on-line precision switch of a DNN model on a device. (a) a device contains two separate models, each trained for the target precision, (b) a device contains a single model, but re-trained on-demand for the target precisions, (c) the proposed dual-precision DNN that contains two different precision modes in a single model, supporting easy switch between different precision modes.

## 2. RELATED WORK
In recent years, a growing number of studies have proposed techniques for reducing the bit precision of DNN models.

One line of studies proposed DNN training methods to minimize accuracy loss under quantization of weights and/or activations. DNNs such as BinaryConnect [4], BNN [7], and XNOR-Net [16] have been proposed to quantize weights/activations to 1-bit. TWN [8,13] added the level zero, and other studies [5,19] have represented parameters as multiple levels expressed as integers using the rounding function. To exploit the bell-shaped distribution of the weights, LogNet [12] represented quantized parameters in the log domain. Another line of studies [6,21] applied re-training after quantization to compensate accuracy drop. As the above quantization methods are tightly coupled with the training process, changing the model precision requires retraining of the model, otherwise the accuracy drop can be significant [3].

To enable precision down-scaling without training, several studies have pro- posed post quantization techniques, which mainly focus on finding the clipping threshold. Sung et al. [18] used L2 error to optimize fixed-point quantization, and Migacz et al. [15] found the minimum value using the KL divergence between the original distribution and the quantized distribution. Banner et al. [1] used Gaussian and Laplacian distribution to model the prior distribution, and Zhao et al. [20] divided the channel of filters to reduce the size of the clip value.

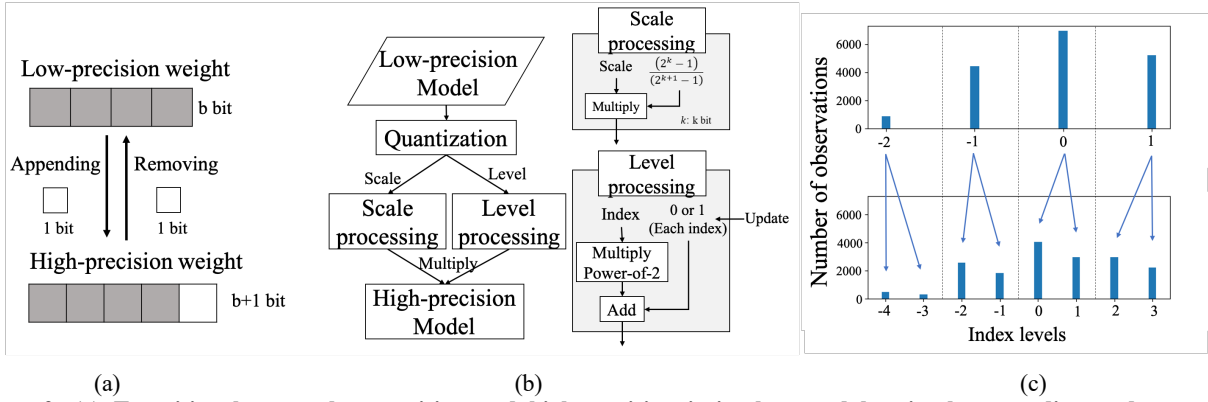

Figure 2. (a) Transition between low-precision and high-precision is implemented by simple appending and removing operations of the up-scaling bit. (b) A flow chart representing the overall process of the proposed training method. (c) Each histogram representing indices and scale corresponding to the bit precision. Top represents the original 2-bit low-precision weights, and bottom represents branched 3-bit high-precision weights.

Most recently, Liu et al. [14] linearly combined multiple low precision weights to approximate full weights. Although the post quantization methods aim to remove the need for re-training after precision down-scaling, down-scaled DNN models generally show lower accuracy than the models originally trained for the same target precision. In addition, to minimize quantization error after post quantization, those methods require finding and tuning hyper-parameters, which can be critical overhead in on-line dynamic switching between different precision modes on the edge devices.

To enable precision switch without additional processing or re-training, this paper proposes dual-precision DNNs that readily include two distinct weight precision modes. Unlike the post quantization methods, the proposed technique achieves comparable accuracy to the models originally trained for the same precision.

## 3. DUAL-PRECISION DEEP NEURAL NETWORK

The proposed method creates dual-precision DNNs by training both the low- precision ($b$ bits) and high-precision ($b$+1 bits) parts of the weights. We propose weight-bit sharing between the two precision modes, where the $b$-bit values of the weights in the low-precision mode are directly used for the $b$ high-order bits of the weights in the high-precision mode (Fig 2(a)). This bit sharing allows efficient transition between the two models during inference, realized with simple appending and removing of the 1 up-scaling bits.

In order to include dual precision modes in each weight element, we propose the training process as a series of shared bit training (optimizing both the low- and high-precision weights) and up-scaling bit training (optimizing the high- precision weights), as shown in (Fig 2(b)). As we train dual-precision DNNs, each level of the $b$-bit low-precision weights are branched out into 2 levels, to make more index values supported by the high-precision weights (Fig 2(c)). The following sections describe detailed quantization and training method for dual-precision DNNs.

### 3.1 Quantization Method

Quantization of DNN weights is a process of mapping the weights into different levels, $\mathbf{I}\in\Re^3$ based on the number of levels and the scale $s\in\Re^1$, which are determined by the target precision. Generally, if two DNN models are individually trained for two different target precision, there will be no dependencies between the weight values of the two models. To integrate two precision modes in a single set of weights, the training method should maintain dependencies between the values in the weights with the two precision modes.

To address this, the proposed training method builds the up-scaled $b$+1-bit model based on the original $b$-bit model, while maintaining the weight levels and values of the $b$ shared bits. This can be accomplished by branching out each of the existing weight levels into 2 levels, as shown in (1).

$$I^{b+1} \leftarrow 2 \cdot I^b + \lambda. \quad (1)$$

In terms of representation, the proposed method builds high-precision weights by shifting the shared bits to the left and appending up-scaling bits at the right (at the least-significant position), as depicted in Fig 2(a). The up-scaling bits can be expressed as level parameters, $\lambda$, which should be trained by the proposed training method. For example, if we up-scale by 1-bit, training process will determine $\lambda$ as 0 or 1.

In case of the scale $s$, several studies [1,9,19,20] have obtained empirical or specific value such as the power-of-2. In this work, we assume the range represented by the original $b$-bit precision weights is equal to that of the up- scaled ($b$+1)-bit precision weights. Therefore, we can define the value of the scale in terms of the interval of the range as shown in (2).

$$(2^b - 1)s_l^b = (2^{b+1} - 1)s_l^{b+1}$$
$$s_l^{b+1} = (2^b - 1)\, s_l^b / (2^{b+1} - 1), \quad (2)$$

Where $\lambda$ is a positive value in the range of $[0, 2^{k-1}]$.

For any quantization method, it is important to properly set the scale according to the range of the weights, to minimize the loss of accuracy. The proposed method specifies the scale value depending on the quantization method – linear quantization.

In linear quantization [5,19], the scale value is defined by dividing the range of weights by the number of levels corresponding bit-width. The weight values are divided by the scale, then applied by the Round function, to be represented as the level index. Therefore, the quantized weights are defined by multiplying the level index and the scale. For linear quantization $q_{\text{linear}}(x; s)$, the levels $\mathbf{I}$ and scale $s$ are obtained by

$$q_{linear}(x; s) := clip(\mathbf{I}; n, p) \cdot s \qquad (3)$$

where $\mathbf{I} = \lceil x/s \rfloor$, $n = -2^{b-1}$, $p = 2^{b-1}-1$, $s = \max(|\mathbf{x}|)/(2^{b-1}-1)$ for signed data.

By applying (1) with the levels $\mathbf{I}$ and scale $s$ of the original precision model, the levels $\mathbf{I}^{b+1}$ and scale $s^{b+1}$ of the upscaled precision model can be obtained. The weights of the up-scaled model are obtained by multiplying $\mathbf{I}^{b+1}$ with $s^{b+1}$.

## 3.2 Training Method

In order to effectively train dual-precision models, both the added level parameters $\lambda$ and the shared $b$-bit low-precision weights should be optimized. To address this motivation, we divide the training process into two phases. In the first phase, hypotheses from both the low-precision and high-precision weights are combined into a single hypothesis, which is then used to update the weights. Weight update is performed in an alternative way, that is, one epoch updates only the $b$-bit weights and the other epoch updates the both low precision $b$-bit weights and high-precision $b+1$-bit weights. The second phase only updates the high-precision weights, based only on the hypothesis from the high-precision weights.

Algorithm 1 and Algorithm 2 describe the detailed process, each of which corresponds to the first and second phase explained above, respectively.

### 3.2.1 Algorithm 1

Algorithm 1 takes the low-precision weights and high-precision weights, with the low-precision weight bits are shared by the both weights. To reflect the errors from the high-precision weights into the low-precision model updates, the final hypothesis h is constructed as the regularized average of the hypotheses of the low-precision weights ($h_{M^b}$) and high-precision weights ($h_{M^{b+1}}$).

$$h = (h_{M^b} + \eta h_{M^{b+1}})/2, \qquad (6)$$

where $\eta \in \Re^1$ is an arbitrary positive real value for regularizing the hypothesis.

The final hypothesis is then compared with the label corresponding to the training data to calculate the gradient updates. Weight update is performed differently in odd and even epochs. In odd epochs, only the shared $b$-bit low-precision parameters are updated. In even epochs, on the other hand, 1-bit level parameters $\lambda$ added in the high-precision weights are updated as well as the shared $b$-bit parameters.

### 3.2.2 Algorithm 2

Algorithm 2 is to train only the up-scaling bits (1-bit) of the high-precision weights, by maintaining the low-precision weights trained during the phase 1. The process in the forward pass is same as in Algorithm 1. The major difference from Algorithm 1 is that Algorithm 2 needs only the high-precision model's hypothesis $h_{M^{b+1}}$ for training, and the trainable parameters are $\lambda$ in the high-precision model.

If the model does not utilize enough levels, the accuracy of the model cannot be high enough. Therefore, to facilitate level branching, we apply index normalization. More specifically, the index parameters are initialized to a normal distribution with a mean of 0 and a standard deviation of 0.3 (m = 0, σ = 0.3), and the range of the updated parameters is scaled as same as the range of the index parameters. By applying this index normalization, we can initialize the index parameters to be distributed evenly so that the high-precision weights can achieve better accuracy.

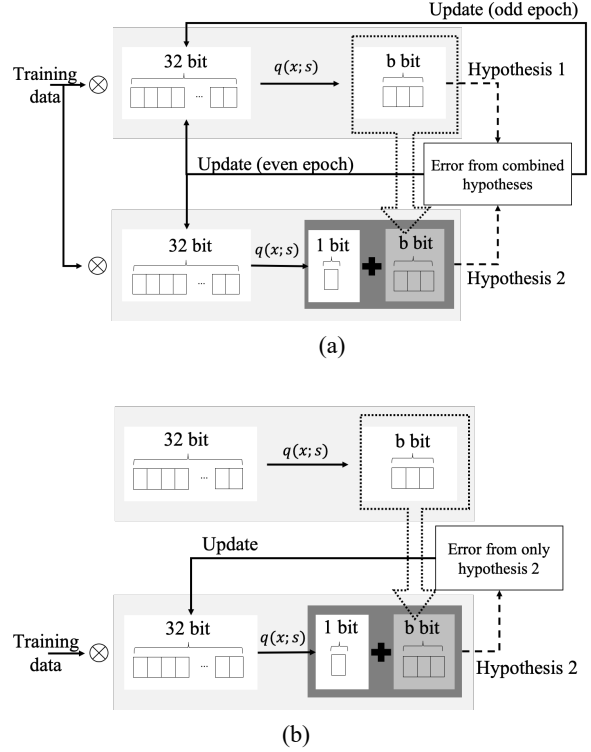

Figure 3. Demonstration of the proposed training algorithm. (a) Algorithm 1, (b) Algorithm 2. Two big squares represent models. The top model is the low-precision model, and the bottom model is the high-precision model.

## 4. EXPERIMENTS

### 4.1 Experimental Configuration

We performed evaluations with AlexNetBN (AlexNet [10] with batch normalization) VGG16 [17], using the CIFAR10 and CIFAR100 datasets [11].

To validate application of linear quantization into the proposed methods, we applied linear quantization for AlexNetBN with CIFAR100, CIFAR10 datasets, and VGG16 with CIFAR100 datasets.

For training, we used batch size of 125, and learning rate of 3e-4, 3e-5, and 4e-3 for the odd and even epochs of the phase 1 and the phase 2, respectively. We used different η depending on the model, but it was set to 0.01 in general. As η gets larger, high-precision model gets better results but low-precision model could not be trained enough. Regarding the transition point from phase 1 and 2, we used 50 epochs.

For CIFAR10, we did not apply any data augmentation, but for CIFAR100, we used data augmentation such as random horizontal flipping and cropping.

### 4.2 Results

Table 1 shows the test accuracy values of the proposed dual-precision DNNs with the two precision modes (n-bit and (n+1)-bit precision). The performance is compared against the baseline method, which trains a dedicated DNN model targeted for each bit precision. For the proposed method, double arrow (↔) indicates that a single DNN model can switch between n-bit to $n+1$ bit weights. As shown in Table 1, the proposed dual-precision DNNs achieve comparable performance to the baseline models dedicated for the target precision. The result shows that the proposed method is effective for linear quantization method.

Table 1. Top-1 accuracy(%) of 1-bit up-scaled dual-precision models and baseline models in CIFAR 10.

| Bit-width | Quantization Method | CIFAR10 AlexNetBN | |
|---|---|---|---|
| | | $n$ bit | $n+1$ bit |
| 1 ↔ 2bit | Baseline | 75.0 | 77.4 |
| | Proposed | 74.5 ↔ | 76.7 |
| 2 ↔ 3bit | Baseline | 77.4 | 79.0 |
| | Proposed | 77.1 ↔ | 79.8 |
| 3 ↔ 4bit | Baseline | 79.0 | 81.5 |
| | Proposed | 79.6 ↔ | 80.7 |
| 4 ↔ 5bit | Baseline | 81.5 | 82.1 |
| | Proposed | 81.4 ↔ | 81.6 |

We also evaluate performance of dual-precision DNNs with VGG16 with CIFAR100 datasets, as shown in Table 2. As in the AlexNetBN with CIFAR10 datasets result shown in Table 1, the proposed DNNs with different model and datasets show similar or better performance compared to the baseline models. The results show that the proposed dual-precision DNN models show no significant accuracy loss, as the models readily include high-precision as well as low-precision weights trained for each target precision.

Fig 4(a) represents the test accuracy during the training process of 2-bit/3- bit dual-precision weights in AlexNetBN model for the CIFAR10 dataset. In this figure, the dotted line indicates the transition of the training phase from phase 1 to phase 2 (from Algorithm 1 to Algorithm 2). During the first training phase with Algorithm 1, there is a fluctuation of the accuracy values of the both 2-bit and 3-bit precision weights. This fluctuation is because of the fact that the shared 2-bit weights are alternately updated in even and odd epochs. The accuracy values of the 2-bit weights and 3-bit weights are close each other because they share the same 2-bit weight values. However, the accuracy of the 3-bit weights is not higher than that of the 2-bit weights yet since the additional 1-bit portion is not fully trained in this phase. In the second phase, on the other hand, the full 3-bit weights are trained while maintaining 2-bit weights. Therefore, the accuracy of the 3-bit weights exceeds that of 2-bit weights, while approaching the accuracy of the baseline 3-bit precision DNN model.

Fig 4(b) represents the maximum number of index levels in all the layers of the same DNN model as in Fig 4(a) during the proposed training process. It shows that level branching according to the added 1-bit occurs right after the transition to the phase 2. This is because the phase used η to more focus on the shared 2-bit weights, creating little influence on index parameters to be updated. In contrast, when the model is train in phase 2, the almost all the levels available for the high-precision bits are used to represent the weight levels, thereby enhancing the accuracy.

Table 2. Top-1 accuracy(%) of 1-bit up-scaled dual-precision models and baseline models in CIFAR100.

| Bit-width | Quantization Method | CIFAR100 | | | |
|---|---|---|---|---|---|
| | | AlexNetBN | | VGG16 | |
| | | $n$ bit | $n+1$ bit | $n$ bit | $n+1$ bit |
| 1 ↔ 2bit | Baseline | 54.8 | 56.1 | 66.3 | 66.7 |
| | Proposed | 54.7 ↔ | 55.7 | 66.3 ↔ | 66.7 |
| 2 ↔ 3bit | Baseline | 56.1 | 58.1 | 66.7 | 67.3 |
| | Proposed | 55.7 ↔ | 57.4 | 66.3 ↔ | 66.8 |
| 3 ↔ 4bit | Baseline | 58.1 | 59.2 | 67.3 | 67.9 |
| | Proposed | 57.9 ↔ | 58.9 | 67.2 ↔ | 68.0 |
| 4 ↔ 5bit | Baseline | 59.2 | 60.7 | 67.9 | 67.9 |
| | Proposed | 59.0 ↔ | 60.3 | 67.8 ↔ | 68.1 |

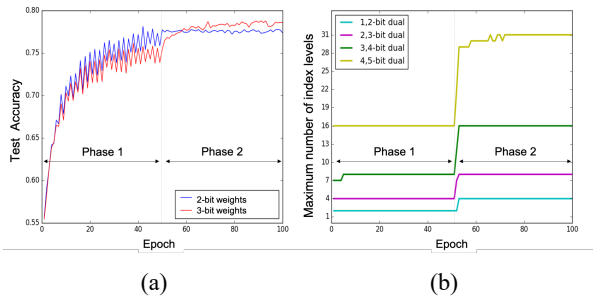

(a)      (b)

Figure 4. (a) accuracy of the low-precision model and the high-precision model during the training process, (b) the maximum number of index levels in each model during training.

## 5. CONCLUSION

This work presents a novel training method for training DNN models targeted for two distinct precision modes in a single model. We show that the proposed method can train a DNN with two precision modes, achieving similar accuracy compared to the baseline models trained for the same precision. The dual-precision DNN allows easy switch between without any retraining, enabling efficient trade-off between the accuracy and compute/memory complexity of the DNN inference.

## 6. REFERENCES

[1] Banner, R., Nahshan, Y., Hoffer, E., Soudry, D.: Aciq: Analytical clipping for integer quantization of neural networks (2018).

[2] Cai, J., Takemoto, M., Nakajo, H.: A deep look into logarithmic quantization of model parameters in neural networks. In: Proceedings of the 10th International Conference on Advances in Information Technology. pp. 1–8 (2018).

[3] Chung, J., Shin, T.: Simplifying deep neural networks for neuromorphic architec- tures. In: 2016 53nd ACM/EDAC/IEEE Design Automation Conference (DAC). pp. 1–6. IEEE (2016).

[4] Courbariaux,M.,Bengio,Y.,David,J.P.:Binaryconnect:Trainin gdeepneuralnet- works with binary weights during propagations. In: Advances in neural information processing systems. pp. 3123–3131 (2015).

[5] Gupta, S., Agrawal, A., Gopalakrishnan, K., Narayanan, P.: Deep learning with limited numerical precision. In: